\documentclass[letterpaper, 12pt]{ewshm}

\usepackage[dvips,final]{graphicx}

\usepackage[table]{xcolor}
\usepackage{makecell}
\usepackage{enumerate}
\usepackage{color}
\usepackage{subfig}
\usepackage{caption}
\usepackage{amsfonts}
\usepackage{amsmath}
\usepackage{amssymb}
\usepackage{times}
\usepackage{cite}
\usepackage{hhline}
\usepackage{compactbib}
\usepackage{graphicx}        
\usepackage{amsmath,amsfonts,amssymb}
\usepackage{geometry}
\usepackage{lineno,hyperref}
\usepackage[labelsep=period]{caption}
\usepackage{siunitx}
\usepackage{tikz}
\usetikzlibrary{bayesnet}
\usetikzlibrary{backgrounds}

\definecolor{halfgreen}{RGB}{0,128,0}
\definecolor{ahsred}{RGB}{192,0,0}

\renewcommand{\thefigure}{\arabic{figure}}
\renewcommand{\thetable}{\Roman{table}}

\newcommand{\beq}{\begin{equation}}
\newcommand{\eeq}{\end{equation}}
\newcommand{\bgqar}{\begin{eqnarray}}
\newcommand{\enqar}{\end{eqnarray}}
\newcommand{\bgqarn}{\begin{eqnarray*}}
\newcommand{\enqarn}{\end{eqnarray*}}
\newcommand{\bgary}{\begin{array}}
\newcommand{\enary}{\end{array}}

\long\def\symbolfootnote[#1]#2{\begingroup%
\def\thefootnote{\fnsymbol{footnote}}\footnote[#1]{#2}\endgroup}

\makeatletter
\renewcommand\@biblabel[1]{#1.}
\makeatother

\begin{document}


\vspace*{4.4cm}

\noindent Title: \textbf{Anomaly Detection in Offshore Wind Turbine Structures using Hierarchical Bayesian Modelling}

\vspace{1.6cm}

\noindent
$
\begin{array}{ll}
\text{Authors}: 
& \text{Simon M. Smith}{^1}  \\ 
& \text{Aidan J. Hughes}{^1}  \\ 
& \text{Tina A. Dardeno}{^1}  \\ 
& \text{Lawrence A. Bull}{^2}  \\ 
& \text{Nikolaos Dervilis}{^1}  \\ 
& \text{Keith Worden}{^1}
\end{array}
$

\newpage


\vspace*{60mm}

\noindent \uppercase{\textbf{ABSTRACT}} \vspace{12pt} 

\noindent Population-based structural health monitoring (PBSHM), aims to share information between members of a population. An offshore wind (OW) farm could be considered as a population of nominally-identical wind-turbine structures. However, benign variations exist among members, such as geometry, sea-bed conditions and temperature differences. These factors could influence structural properties and therefore the dynamic response, making it more difficult to detect structural problems via traditional SHM techniques.


This paper explores the use of a hierarchical Bayesian model to infer expected soil stiffness distributions at both population and local levels, as a basis to perform anomaly detection, in the form of scour, for new and existing turbines. To do this, observations of natural frequency will be generated as though they are from a small population of wind turbines. Differences between individual observations will be introduced by postulating distributions over the soil stiffness and measurement noise, as well as reducing soil depth (to represent scour), in the case of anomaly detection.


\let\thefootnote\relax\footnotetext{\hspace*{-7mm} Simon Smith, PhD Student, Email: ssmith8@sheffield.ac.uk.\ \textsuperscript{1}{Dynamics Research Group, Department of Mechanical Engineering, The University of Sheffield, Western Bank, Sheffield, S10 2TN, UK}; \textsuperscript{2}{Computational Statistics and Machine Learning Group, Department of Engineering, University of Cambridge, Cambridge, CB3 0FA, UK}}


\vspace{24pt} 
\noindent \uppercase{\textbf{INTRODUCTION}}  \vspace{12pt} 


\noindent The design of offshore wind (OW) turbine towers and foundations is driven by fatigue and extreme loading concerns~\cite{noauthor_guide_2019}. To minimise fatigue damage, the resonance frequencies of the dynamic structure must be avoided, to prevent amplification of the response to external forcing~\cite{andersen_natural_2012}.
These forces predominantly come in the form of wind and wave loading in the natural environment, and forces from the rotation of the blades, at both the rotational speed (1P) and three times the rotational speed (3P). Excitation from the 1P frequency is associated with mass imbalance in the blades; with the 3P frequency it is associated with the blade-passing frequency of the tower, causing a shadow effect~\cite{bhattacharya_physical_2021}. 
Typically, the first natural frequencies of the structures are designed to lie between the 1P and 3P frequencies, as a compromise between steering clear of environmental loading, and the cost of additional structural material for stiffness~\cite{bhattacharya_challenges_2014}. Careful consideration must therefore come in the design stage of wind-turbine foundations, to avoid resonance frequencies coinciding with the frequencies of the sources of excitation.


An OW farm could be considered as a homogeneous population of wind-turbine structures. However, from the point of view of their dynamic response, variations exist between members of the population; this includes possible variations in the soil profile across the wind farm, influencing the effective stiffness of the foundation support. Furthermore, perhaps more benign environmental variations also exist, such as in temperature, wave and wind conditions or physical geometry from manufacturing tolerances. According to a study by Sørum et al.~\cite{sorum_fatigue_2022}, dynamic response uncertainties related to wind conditions dominate in the tower top, while uncertainties in the wave and soil models dominate in the tower base and monopile.
Given the relatively-narrow frequency band in which the first natural frequency should lie, minimising uncertainty in the modelling of dynamic response is important to reduce the likelihood of fatigue damage in operation. Furthermore, from the point of view of performing SHM, any uncertainties that do remain in the expected dynamic response may make identification of damage that influences the structural properties more challenging. For example, a significant problem in the operation of wind farms is the scouring of sediment around monopile foundations at the interface between seawater and sea bed; this in effect reduces the embedded depth of the monopile, which, in turn, reduces the stiffness of the support~\cite{prendergast_investigation_2013, prendergast_investigation_2015}. As a consequence, the dynamic response of the entire structure is affected. However, if the reduction in natural frequency is within the expected variance in the dynamic response of an individual structure, it may go undetected.

This work adopts a hierarchical Bayesian modelling approach (recently used in the application of PBSHM~\cite{bull_hierarchical_2022, dardeno_population_2023}), to develop population and individual turbine-level distributions for the natural frequency of the first bending mode of the structures, which share a level of information between each other. Observations of the natural frequencies are generated via a finite element model of the structure, using soil stiffnesses from predefined distributions as inputs. Approximations of these stiffness distributions are then recovered using the observations of natural frequency within the hierarchical model, and are used as a basis for anomaly detection for new natural frequency measurements.


\vspace{24pt}

\noindent \uppercase{\textbf{FE model construction}} \vspace{12pt}

\noindent A finite element (FE) model was constructed to model the dynamic response of a wind-turbine structure, with geometry and material properties based on the widely studied NREL 5MW reference turbine~\cite{jonkman_definition_2009}. Timoshenko beam elements~\cite{davis_timoshenko_1972} were used in a tubular construction for the monopile and tower, with a lumped mass placed at the top to represent the mass of the nacelle, rotor and blades. For simplicity, the transition piece was not modelled (normally used to mate the monopile to the tower), with the cross-sections of the monopile and tower joined and effectively modelled as a single beam.

\begin{figure}[!b] 
    \centering{\includegraphics[width=0.58\columnwidth]{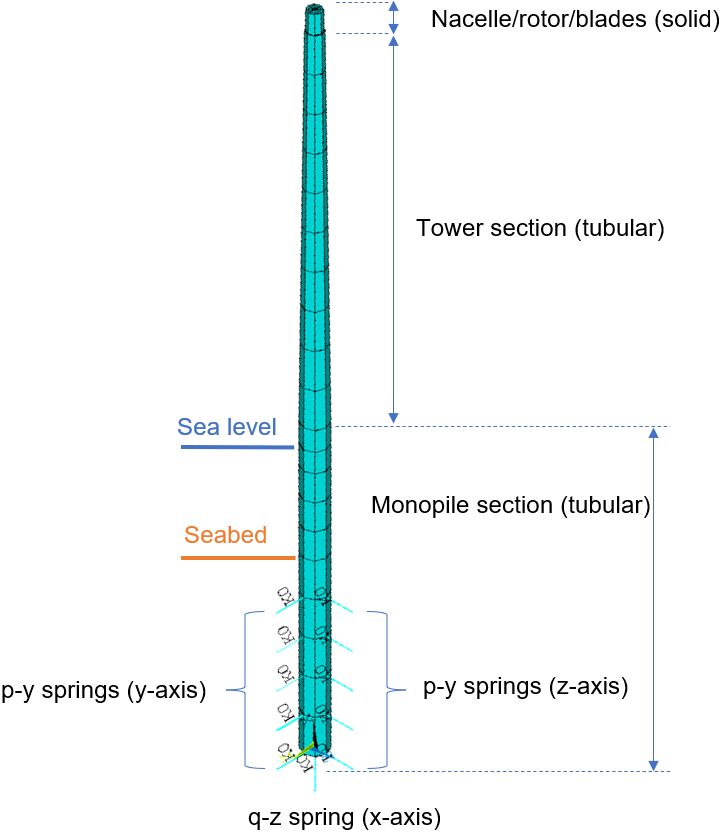}}
      \caption{FE model construction in relation to the offshore environment.}
  \label{fig:fe_model}%
  \end{figure}

A Winkler foundation was adopted to model the soil structure interaction (SSI), using a series of springs to resist lateral movement (known as \textit{p-y} springs), and a spring at the base of the foundation to resist axial movement (known as a \textit{q-z} spring)~\cite{winkler_lehre_1868}. This method was chosen, as the concept of scour can be intuitively modelled by removing springs near the surface of the soil, and reducing the effective depth of the remaining springs. It is also relatively simple to implement, and is the currently-used method by the OW industry, in the \textit{DNV-OS-J101} standard for the design of OW turbine structures~\cite{noauthor_dnv-os-j101_2014}. Figure \ref{fig:fe_model} shows the setup of the FE model, and how it relates to the environment in which it operates.


In this initial approach, a linear spring stiffness was used that did not vary with soil depth for simplicity and to demonstrate the methodology. To model the influence of seawater on the dynamic response of the overall structure, an added-mass technique was used as in~\cite{zuo_dynamic_2018}; this arises from the submerged body being able to impart acceleration to the surrounding fluid~\cite{bi_using_2016}. Additional components on the steel tower were also accounted for, such as bolts and flanges, by modelling the density of the steel tower as 8500 $kg/m^3$ - as in NREL's 5MW reference turbine~\cite{jonkman_definition_2009}. 

To give confidence in the FE model results and to be comparable to a known structure, the bending-mode natural frequencies of the tower and nacelle section, fixed at the base (i.e.\ not including the SSI), were first compared against the NREL 5MW reference turbine~\cite{jonkman_definition_2009} and were found to be consistent. Secondly, including the monopile and Winkler foundation model, the soil stiffness was then tuned to be similar to the bending-mode natural frequencies determined in Zuo et al.~\cite{zuo_dynamic_2018}.


\vspace{24pt}
\noindent \uppercase{\textbf{Hierarchical Bayesian Model}} \vspace{12pt}

\noindent The explanation here follows the description provided by Bull et al. \cite{bull_hierarchical_2022}. Consider structural data, recorded from a population of $K$ similar wind turbines in comparable soil conditions. The population data can be denoted,

\begin{equation}\label{eq:data}
  \left\{\mathbf{x}_k, \mathbf{y}_k\right\}_{k=1}^K=\left\{\left\{x_{i k}, y_{i k}\right\}_{i=1}^{N_k}\right\}_{k=1}^K
\end{equation}
where $\mathbf{y}_{k}$ is a target response vector for inputs $\mathbf{x}_{k}$ and $\{x_{ik},y_{ik}\}$ are the $i^{\text{th}}$ pair of observations for turbine $k$. There are $N_{k}$ observations for each turbine and thus $\Sigma^{K}_{k=1}N_k$ observations in total. 



Models that include both shared parameters learnt at the population level (often referred to as fixed effects), and random effects which vary between groups/subfleets (\textit{K}), are known as partially-pooled hierarchical models. In contrast, a no-pooling approach is where each sub-fleet has an independent model, whereby no statistical strength is shared between domains.
In the OW setting, whilst some older wind turbines may have a rich history of data, newer wind turbines may not, and so independent models may lead to unreliable predictions. On the other end of the spectrum, a complete-pooling approach considers all population data from a single source~\cite{dardeno_population_2023}; this may lead to poor generalisation, particularly when there are significant differences between individual or groups of wind turbines (wind farms). Hierarchical (partial-pooling) models represent a middle-ground which can be used to learn separate models for each group while encouraging task parameters to be correlated. 

In this work, the hierarchical model was used to learn a global distribution over the turbine soil stiffnesses and assumed the soil stiffness associated with each foundation was a sample from this shared global distribution; these values were used as inputs to a predetermined polynomial function, that mapped foundation soil stiffness to the natural frequency of the first bending mode. The polynomial function is a surrogate for the mapping of the FE foundation model from which it was fitted for efficiency of computation during the sampling; this was performed using the Monte Carlo Markov Chain (MCMC) method, via the no U-turn (NUTS) implementation of Hamiltonian Monte Carlo (HMC)~\cite{homan_no-u-turn_2014}.


The following explanation describes the setup of the model. The likelihood for the model is,
\begin{equation} \label{eq:likelihood}
  \left\{\left\{\omega_{ik}\right\}_{i=1}^{N_k}\right\}_{k=1}^K {\sim} \mathrm{Normal}\left(f({\text{exp}(s_{k})}),{\gamma}^2\right)
\end{equation}
where $\omega_{ik}$ is the first natural frequency of the $i^{th}$ observation and the $k^{th}$ turbine, $f$ the polynomial function approximating the FE model, and $\gamma$ representing natural frequency measurement error. Following the Bayesian methodology, one can set prior distributions over the stiffnesses, $s_k$, for each turbine:
\begin{equation} \label{eq:s_k}
  \left\{{s}_k\right\}_{k=1}^K {\sim} \mathrm{Normal}\left(ln\left(\frac{{{{\mu}_s}^2}}{\sqrt{{{{\mu}_s}^2}+{{{\sigma}_s}^2}}}\right),ln\left(1+\frac{{{\sigma}_s}^2}{{{\mu}_s}^2}\right)\right)
\end{equation}
\begin{equation} \label{eq:mu_s}
  {\mu}_s \sim \mathrm{Normal}\left({{\mu}}_{\mu},{\sigma}_{\mu}^2\right)
\end{equation}
\begin{equation} \label{eq:sig_s}
  {\sigma}_s \sim \mathrm{HalfCauchy}\left(0,{{\beta}_{\sigma}}\right)
\end{equation}
Note that the mean and standard deviation in equation $\left(\ref{eq:s_k}\right)$ and the exponent in equation $\left(\ref{eq:likelihood}\right)$ are formulated in this way to avoid negative stiffnesses being possible, whilst maintaining intuitive understanding of the individual parameters. Equation $\left(\ref{eq:mu_s}\right)$ shows that the prior expectation of the population-level stiffness is also normally distributed, with mean ${\mu}_{\mu}$  and standard deviation ${\sigma_{\mu}}$. Equation $\left(\ref{eq:sig_s}\right)$ shows that the prior standard deviation $\sigma_{s}$ of the population-level stiffness is HalfCauchy distributed with scale parameter ${\beta_{\sigma}}$,

\begin{equation} \label{eq:noise}
  {\gamma} {\sim} \mathrm{HalfCauchy}\left(0,{\beta_{\gamma}}\right)
\end{equation}

\begin{figure}[b]
  \centering
  \begin{tikzpicture}
    \node[obs]                              (w) {$\omega_{ik}$};
    \node[latent, left=1.5cm of w] (s_k) {$s_k$};
    \node[latent, above=0.5cm of s_k, xshift=-1.5cm] (mu_s) {$\mu_{s}$};
    \node[latent, below=0.5cm of s_k, xshift=-1.5cm] (sig_s) {$\sigma_{s}$};
    \node[draw=none, above=0.5cm of mu_s, xshift=-1.5cm] (mu_mu) {$\mu_{\mu}$};
    \node[draw=none, below=0.5cm of mu_s, xshift=-1.5cm] (sig_mu) {$\sigma_{\mu}$};
    \node[draw=none, left=0.9cm of sig_s] (beta_sig) {$\beta_{\sigma}$};
    \node[latent, right=1cm of w] (noise) {$\gamma$};
    \node[draw=none, right=1cm of noise] (hypernoise) {$\beta_{\gamma}$};

    \edge {s_k,noise} {w} ; %
    \edge {mu_s,sig_s} {s_k} ; %
    \edge {mu_mu,sig_mu} {mu_s} ; %
    \edge {beta_sig} {sig_s} ; %
    \edge {hypernoise} {noise} ; %

    \plate {N} {(w)} {$i\in1:N$} ;
    \plate {k} {(w)(s_k)(N.north west)(N.south east)} {$k\in1:K$} ;

  \end{tikzpicture}
  \caption{A graphical model representing the hierarchical model with partial pooling.}
  \label{fig:hierarchical_model}
\end{figure}
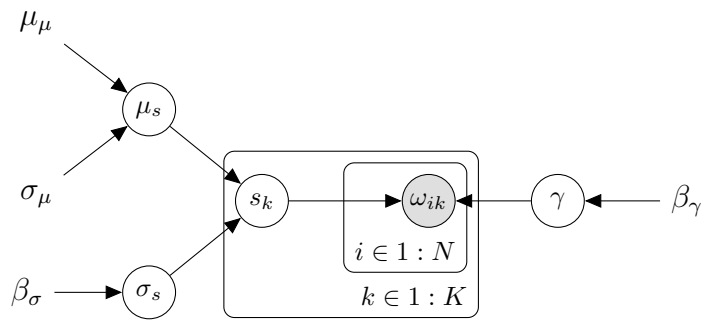

Finally, the standard deviation of ${\omega}_{ik}$, ${\gamma}$, is HalfCauchy distributed, with scale parameter ${\beta}_{\gamma}$. Gelman et al.~\cite{gelman_prior_2006} recommends the use of HalfCauchy distributions for population variances, as their heavy tails bring a robustness against outliers to the model, as well as efficiency during the inference and sampling process. A graphical model depicting the hierarchical structure can be seen in Figure \ref{fig:hierarchical_model}. Latent and observed variables are depicted as unshaded and shaded circled nodes respectively, with the plates indicating multiple instances of their containing nodes. The uncircled parameters are the constants used in the prior distributions.





\vspace{24pt}
\noindent \uppercase{\textbf{Dataset Generation}} \vspace{12pt}

\begin{figure}[t] 
  \centering{\includegraphics[width=0.8\columnwidth]{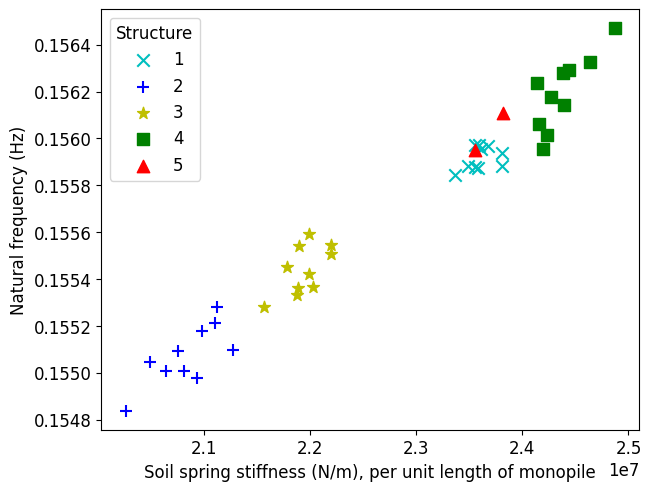}}
    \caption{Generated natural frequency observations for five turbine structures.}
\label{fig:gen_data}%
\end{figure}

\noindent In this initial approach, the dataset (i.e. observations of the first natural frequency) was generated in conjunction with the FE model of the wind-turbine foundation. Distributions were placed over global stiffness expectation and variance, which were used to sample $K$ local stiffness expectations and variances. For each $k$, $N$ stiffness realisations were then drawn, and passed through the FE model to generate natural frequency observations. Finally, these observations were corrupted by independently sampled Gaussian noise from a normal distribution, with mean zero and standard deviation equal to \num{e-4}. The first four of five structures were given 10 observations, whilst the final structure was given just two, to emulate a new structure with limited data. This created an imbalanced dataset with a total of 42 observations. Figure ~\ref{fig:gen_data} shows these observations used in the analyses, where the red triangles show the fifth (new) structure with limited data.

\vspace{24pt}
\noindent \uppercase{\textbf{Preliminary results and Discussion}} \vspace{12pt} 

\begin{figure}[t] 
  \centering{\includegraphics[width=0.8\columnwidth]{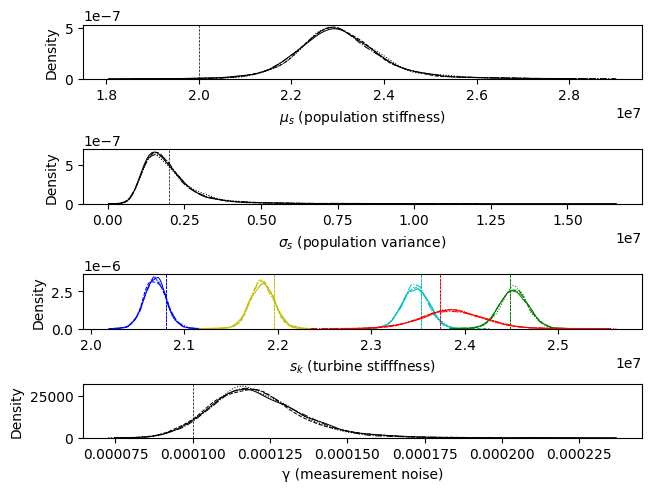}}
    \caption{Density plots showing the posterior samples for each chain, for the learned parameters. Vertical dashed lines indicate the expected values used in generating the data. Black lines indicate population-level parameter posteriors while the coloured lines (for $s_k$) represent each turbine $k$ in $1:K$.}
\label{fig:post_trace}%
\end{figure}

\noindent As is typical, due to the high computational demand of the FE model, it was infeasible to incorporate it directly as the mean function within the probabilistic model. Instead, a surrogate 5${^{th}}$-order polynomial function was used that closely matched the relationship between the input soil spring stiffness per unit length, and the natural frequency of the first bending mode. Specifically, this was fitted to the range of soil spring stiffnesses between \num{e7} and \num{5e7} $N/m$ (per unit length), which was deemed a reasonable magnitude of range at this stage. The MCMC sampling was then carried out using four chains, each with 2000 warm-up samples, followed by a further 2000 samples. The warm-up samples were discarded to diminish the influence of starting values~\cite{gelman_bayesian_2013}. Density plots for the further 2000 samples for each chain are shown in Figure~\ref{fig:post_trace}, representing the inferred distributions of the parameters. The vertical dashed lines mark the true expected values of the distributions used in generating the data. 

It can be seen that the sampling appears to have stabilised with all four chains (shown by each line style) producing very similar results across all parameters. Given that the higher level model is learnt with only observations at the bottom level, the modes of the posterior distributions are relatively close to their expected values. The variance of soil spring stiffness for the fifth structure (shown in red) is also comparably larger, as expected, due to fewer data points to learn from. In all cases, the variances of the distributions are considerably smaller than those in the prior beliefs that were placed on each parameter. These observations give confidence that the algorithm is working as expected. On real wind turbine structural data, this could give greater confidence in soil parameters that could aid in the design process of wind-turbine structures in comparable ground conditions.

\begin{figure}[t] 
  \centering{\includegraphics[width=0.8\columnwidth]{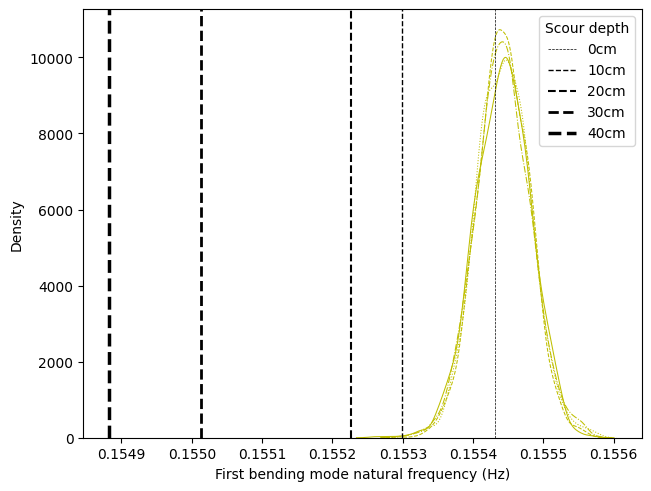}}
    \caption{Average natural frequency of five samples for increasing levels of scour depth, in comparison to the posterior distribution of natural frequency for the third structure.}
\label{fig:scour}%
\end{figure}

As an example of how the posterior distributions of the stiffness could be used for anomaly detection, Figure ~\ref{fig:scour} shows the posterior natural frequency for the third structure in yellow, computed from the stiffness samples of the posterior via the FE model. 
The vertical dashed lines represent the averages of five samples including five different levels of scour in 10 centimetre increments, modelled by reducing the length of the monopile which the springs acted upon. In this case, it can be seen that the scour depths of 30 cm and 40 cm result in natural frequencies in regions of very low posterior density. In a real scenario, observing frequencies in an equivalent region could give an indication that there is enough scour present to reduce the natural frequency below the expected range for the soil conditions. It is important to note, however, that at this stage other sources of uncertainty, such as wind and wave loading, and more realistic non-linear modelling of the soil stiffness have not yet been incorporated.
\vspace{24pt}

\noindent \uppercase{\textbf{CONCLUDING REMARKS}} \vspace{12pt}

\noindent A hierarchical Bayesian model has been constructed to model the expected first bending natural frequency of a small population of wind-turbine structures, considering the uncertainty in soil stiffness. This method shows that \textit{a priori} beliefs on the uncertainty in soil stiffness can be updated based on observations of natural frequency, which could help inform future structural design in areas with comparable soil conditions, or be used as a basis for more robust anomaly detection. Future work should consider wider sources of uncertainty for natural frequency measurements. These include modelling the soil stiffness as non-linear, and increasing the stiffness with depth. The stochastic nature of wind and wave loading could also be explored, which influences structural deflection, and therefore stiffness in the non-linear case.



\vspace{36pt}
\noindent \uppercase{\textbf{Acknowledgments}} \vspace{12pt}


\noindent The authors gratefully acknowledge the support of the UK Engineering and Physical
Sciences Research Council (EPSRC) and the Natural Environment Research council (NERC), via grant references EP/W005816/1, EP/R003645/1 and EP/S023763/1. For the purpose of open access, the author(s) has/have applied a Creative Commons Attribution (CC BY) licence to any Author Accepted Manuscript version arising.

\vspace{24pt}

\small 

\bibliographystyle{iwshm}
\bibliography{IWSHM2023}

\begin{thebibliography}{10}
\newcommand{\enquote}[1]{``#1''}
\expandafter\ifx\csname urlstyle\endcsname\relax
  \providecommand{\doi}[1]{doi:\discretionary{}{}{}#1}\else
  \providecommand{\doi}{doi:\discretionary{}{}{}\begingroup
  \urlstyle{rm}\Url}\fi

\bibitem{noauthor_guide_2019}
BVGA. 2019. \enquote{Guide to an offshore wind farm,} Tech. rep.

\bibitem{andersen_natural_2012}
Andersen, L.~V., M.~J. Vahdatirad, M.~T. Sichani, and J.~D. Sørensen. 2012.
  \enquote{Natural frequencies of wind turbines on monopile foundations in
  clayey soils—{A} probabilistic approach,} \emph{Computers and Geotechnics},
  43:1--11.

\bibitem{bhattacharya_physical_2021}
Bhattacharya, S., D.~Lombardi, S.~Amani, M.~Aleem, G.~Prakhya, S.~Adhikari,
  A.~Aliyu, N.~Alexander, Y.~Wang, L.~Cui, S.~Jalbi, V.~Pakrashi, W.~Li,
  J.~Mendoza, and N.~Vimalan. 2021. \enquote{Physical {Modelling} of {Offshore}
  {Wind} {Turbine} {Foundations} for {TRL} ({Technology} {Readiness} {Level})
  {Studies},} \emph{Journal of Marine Science and Engineering}, 9(6):589.

\bibitem{bhattacharya_challenges_2014}
Bhattacharya, S. 2014. \enquote{Challenges in {Design} of {Foundations} for
  {Offshore} {Wind} {Turbines},} \emph{Institution of Engineering and
  Technology}, 1.

\bibitem{sorum_fatigue_2022}
Sørum, S.~H., G.~Katsikogiannis, E.~E. Bachynski-Polić, J.~Amdahl, A.~M.
  Page, and R.~T. Klinkvort. 2022. \enquote{Fatigue design sensitivities of
  large monopile offshore wind turbines,} \emph{Wind Energy},
  25(10):1684--1709.

\bibitem{prendergast_investigation_2013}
Prendergast, L.~J., D.~Hester, K.~Gavin, and J.~J. O’Sullivan. 2013.
  \enquote{An investigation of the changes in the natural frequency of a pile
  affected by scour,} \emph{Journal of Sound and Vibration},
  332(25):6685--6702.

\bibitem{prendergast_investigation_2015}
Prendergast, L.~J., K.~Gavin, and P.~Doherty. 2015. \enquote{An investigation
  into the effect of scour on the natural frequency of an offshore wind
  turbine,} \emph{Ocean Engineering}, 101:1--11.

\bibitem{bull_hierarchical_2022}
Bull, L.~A., D.~Di~Francesco, M.~Dhada, O.~Steinert, T.~Lindgren, A.~K.
  Parlikad, A.~B. Duncan, and M.~Girolami. 2022. \enquote{Hierarchical
  {Bayesian} {Modelling} for {Knowledge} {Transfer} {Across} {Engineering}
  {Fleets} via {Multitask} {Learning},} \emph{Computer-Aided Civil and
  Infrastructure Engineering}:mice.12901.

\bibitem{dardeno_population_2023}
Dardeno, T.~A., L.~A. Bull, N.~Dervilis, and K.~Worden. 2023. \enquote{A
  population form via hierarchical {Bayesian} modelling of the {FRF},} in
  \emph{Proceedings of the 41st IMAC, A Conference and Exposition on Structural
  Dynamics}, Austin, USA.

\bibitem{jonkman_definition_2009}
Jonkman, J., S.~Butterfield, W.~Musial, and G.~Scott. 2009. \enquote{Definition
  of a 5-MW Reference Wind Turbine for Offshore System Development,} Tech. Rep.
  NREL/TP-500-38060, 947422.

\bibitem{davis_timoshenko_1972}
Davis, R., R.~D. Henshell, and G.~B. Warburton. 1972. \enquote{A {Timoshenko}
  beam element,} \emph{Journal of Sound and Vibration}, 22(4):475--487.

\bibitem{winkler_lehre_1868}
Winkler, E. 1868. \emph{Die {Lehre} von der {Elastizität} und {Festigkeit} mit
  besonderer {Rücksicht} auf ihre {Anwendung} in der {Technik}: für
  polytechnische {Schulen}, {Bauakademien}, {Ingenieure}, {Maschinenbauer},
  {Architecten}, etc}, no. v. 1 in Die {Lehre} von der {Elastizität} und
  {Festigkeit} mit besonderer {Rücksicht} auf ihre {Anwendung} in der
  {Technik}: für polytechnische {Schulen}, {Bauakademien}, {Ingenieure},
  {Maschinenbauer}, {Architecten}, etc, Dominicius.

\bibitem{noauthor_dnv-os-j101_2014}
DNV. 2014. \enquote{DNV-OS-J101 Design of Offshore Wind Turbine Structures,}
  Tech. rep.

\bibitem{zuo_dynamic_2018}
Zuo, H., K.~Bi, and H.~Hao. 2018. \enquote{Dynamic analyses of operating
  offshore wind turbines including soil-structure interaction,}
  \emph{Engineering Structures}, 157:42--62.

\bibitem{bi_using_2016}
Bi, K. and H.~Hao. 2016. \enquote{Using pipe-in-pipe systems for subsea
  pipeline vibration control,} \emph{Engineering Structures}, 109:75--84.

\bibitem{homan_no-u-turn_2014}
Hoﬀman, M.~D. and A.~Gelman. 2014. \enquote{The {No}-{U}-{Turn} {Sampler}:
  {Adaptively} {Setting} {Path} {Lengths} in {Hamiltonian} {Monte} {Carlo},}
  \emph{Journal of Machine Learning Research}, 15:1593--1623.

\bibitem{gelman_prior_2006}
Gelman, A. 2006. \enquote{Prior distributions for variance parameters in
  hierarchical models (comment on article by {Browne} and {Draper}),}
  \emph{Bayesian Analysis}, 1(3):515--534.

\bibitem{gelman_bayesian_2013}
Gelman, A., J.~Carlin, H.~Stern, D.~Dunson, A.~Vehtari, and D.~Rubin. 2013.
  \emph{Bayesian {Data} {Analysis}, {Third} {Edition}}, Chapman \& {Hall}/{CRC}
  {Texts} in {Statistical} {Science}, Taylor \& Francis.

\end{thebibliography}


\end{document}